\pdfoutput=1

\documentclass[11pt]{article}

\usepackage[]{EMNLP2022}

\usepackage{times}
\usepackage{latexsym}

\usepackage[T1]{fontenc}

\usepackage[utf8]{inputenc}

\usepackage{microtype}
\usepackage{graphicx}
\usepackage{amsmath}
\usepackage{times}
\usepackage{latexsym}
\usepackage{natbib}
\usepackage{enumerate}
\usepackage{amssymb}
\usepackage{multirow}
\usepackage{multicol}
\usepackage{enumitem}
\usepackage{booktabs}
\usepackage{enumitem}
\usepackage{balance}
\usepackage{natbib}
\usepackage{amssymb}
\usepackage{pifont}
\newcommand{\xmark}{\ding{55}}
\title{PAR: Political Actor Representation Learning with \\ Social Context and Expert Knowledge}

\author{Shangbin Feng$^{\clubsuit\spadesuit}$ \: \: 
Zhaoxuan Tan$^{\clubsuit}$ \: \: 
Zilong Chen$^{\clubsuit\heartsuit}$ \: \:  Ningnan Wang$^{\clubsuit}$ \: \: \\ \bf
Peisheng Yu$^{\clubsuit\diamondsuit}$ \: \: Qinghua Zheng$^{\clubsuit}$ \: \: Xiaojun Chang$^{\bigcirc}$ \: \:  Minnan Luo$^{\clubsuit}$\thanks{\ \ Corresponding author.}
\\
Xi’an Jiaotong University$^{\clubsuit}$, University of Washington$^{\spadesuit}$, Tsinghua University$^{\heartsuit}$ \\
University of California San Diego$^{\diamondsuit}$, University of Technology Sydney$^{\bigcirc}$ \\
contact: \href{mailto:shangbin@cs.washington.edu}{\texttt{shangbin@cs.washington.edu}}
}

\begin{document}
\maketitle
\begin{abstract}
Modeling the ideological perspectives of political actors is an essential task in computational political science with applications in many downstream tasks. Existing approaches are generally limited to textual data and voting records, while they neglect the rich social context and valuable expert knowledge for holistic ideological analysis. In this paper, we propose \textbf{PAR}, a \textbf{P}olitical \textbf{A}ctor \textbf{R}epresentation learning framework that jointly leverages social context and expert knowledge. Specifically, we retrieve and extract factual statements about legislators to leverage social context information. We then construct a heterogeneous information network to incorporate social context and use relational graph neural networks to learn legislator representations. Finally, we train PAR with three objectives to align representation learning with expert knowledge, model ideological stance consistency, and simulate the echo chamber phenomenon. Extensive experiments demonstrate that PAR is better at augmenting political text understanding and successfully advances the state-of-the-art in political perspective detection and roll call vote prediction. Further analysis proves that PAR learns representations that reflect the political reality and provide new insights into political behavior.
\end{abstract}

\section{Introduction}
Modeling the perspectives of political actors has applications in various downstream tasks such as roll call vote prediction ~\citep{mou2021align} and political perspective detection ~\citep{feng2021knowledge}. Existing approaches generally focus on voting records or textual information of political actors to induce their stances. Ideal point model ~\citep{clinton2004statistical} is one of the most widely used approach for vote-based analysis, while later works enhance the ideal point model ~\citep{kraft2016embedding,gu2014topic,gerrish2011predicting} and yield promising results on the task of roll call vote prediction. For text-based methods, text analysis techniques are combined with textual information in social media posts ~\citep{li2019encoding}, Wikipedia pages ~\citep{feng2021knowledge}, legislative text ~\citep{mou2021align} and news articles ~\citep{li2021using} to enrich the perspective analysis process.

\begin{figure}
    \centering
    \includegraphics[width=1.0\linewidth]{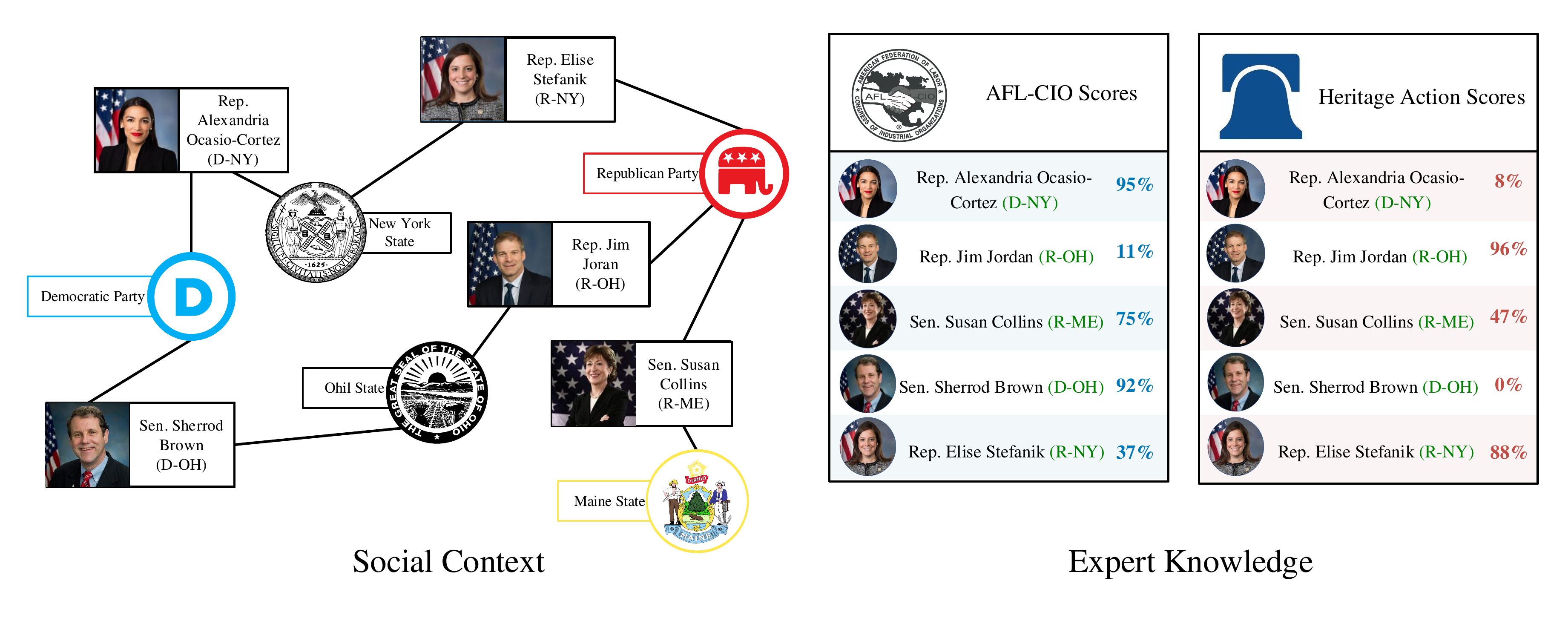}
    \caption{Social context information and political expert knowledge that helps to model political actors.}
    \label{fig:teaser}
\end{figure}

However, existing methods fail to incorporate the rich social context and valuable expert knowledge of political actors. As illustrated in Figure \ref{fig:teaser}, social context information such as home state and party affiliation serves as background knowledge and helps connect different political actors \citep{yang2021joint}. These social context facts about political actors also differentiate them and indicate their ideological stances. In addition, expert knowledge from political think tanks provides valuable insights and helps to anchor the perspective analysis process. As a result, political actor representation learning should be guided by domain expertise to facilitate downstream tasks in computational political science. That being said, social context and expert knowledge should be incorporated in modeling legislators to ensure a holistic evaluation.


In light of these challenges, we propose PAR, a legislator representation learning framework that jointly leverages social context and expert knowledge. We firstly collect a dataset of political actors by retrieving and extracting social context information from their Wikipedia homepages and adapting expert knowledge from two political think tanks AFL-CIO\footnote{\url{https://aflcio.org/}} and Heritage Action\footnote{\url{https://heritageaction.com/}}. After that, we construct a heterogeneous information network to model social context information and adopt relational graph neural networks for representation learning. Finally, we train the framework with three training objectives to leverage expert knowledge, model social and political phenomena, and learn representations of political actors in the process. We evaluate PAR on two computational political science tasks and examine the political behaviour of learned representations. Our main contributions are summarized as follows:

\begin{itemize}[leftmargin=*]
    \item We propose PAR, a graph-based approach to learn legislator representations with three training objectives, which aligns representation learning with expert knowledge, ensures stance consistency, and models the echo chamber phenomenon in socio-economic systems.
    \item Extensive experiments demonstrate that PAR advances the state-of-the-art on two computational political science tasks, namely political perspective detection and roll call vote prediction.
    \item Further analysis shows that PAR learns representations that reflect the ideological preferences and political behavior of political actors such as legislators, governors, and states. In addition, PAR provides interesting insights into the contemporary political reality.
    
\end{itemize}



\section{Related Work}
The ideological perspectives of political actors play an essential role in their individual behavior and adds up to influence the overall legislative process \citep{freeden2006ideology,bamman2012gender,wilkerson2017large}. Political scientists first explored to quantitatively model political actors based on their voting behaviour. Ideal point model~\citep{clinton2004statistical} is one of the earliest approach to leverage voting records to analyze their perspectives. It projects political actors and legislation onto one-dimensional spaces and measure distances. Many works later extended the ideal point model. \citet{gerrish2011predicting} leverages bill content to infer legislator perspectives. \citet{gu2014topic} introduces topic factorization to model voting behaviour on different issues to establish a fine-grained approach. \citet{kraft2016embedding} models legislators with multidimensional ideal vectors to analyze voting records. \citet{mohammad2017stance} and \citet{kuccuk2018stance} use SVM and handcrafted text features to augment the analysis.


In addition to voting, textual data such as speeches and public statements are also leveraged to model political perspectives \cite{evans2007recounting, thomas2006get,hasan2013stance, zhang-etal-2022-kcd, sinno-etal-2022-political, liu-etal-2022-politics, davoodi-etal-2022-modeling, alkiek-etal-2022-classification, dayanik-etal-2022-improving, pujari-goldwasser-2021-understanding, villegas2021analyzing, li-etal-2021-p, sen-etal-2020-reliability, baly-etal-2020-detect}. \citet{volkova2014inferring} leverage message streams to inference users political preference. \citet{johnson2016identifying} propose to better understand political stances by analyzing politicians' tweet and their temporal activities. \citet{li2019encoding} propose to analyze Twitter posts to better understand news stances.
\citet{prakash2020incorporating} and \citet{kawintiranon2021knowledge} leverage pre-trained language model for stance prediction. 
\citet{augenstein2016stance} uses conditional LSTM to encode tweets for stance detection. Furthermore, CNN \cite{wei2016pkudblab} and hierarchical attention networks \cite{sun2018stance} are adopted for stance detection.
\citet{darwish2020unsupervised} proposes an unsupervised framework for user representation learning and stance detection.
\citet{yang2021joint} proposes to jointly model legislators and legislations for vote prediction.
\citet{feng2021knowledge} introduces Wikipedia corpus and constructs knowledge graphs to facilitate perspective detection. \citet{mou2021align} proposes to leverage tweets, hashtags and legislative text to grasp the full picture of the political discourse.
\citet{li2021using} designs pre-training tasks with social and linguistic information to augment political analysis.
However, these vote and text-based methods fail to leverage the rich social context of political actors and the valuable expert knowledge of political think tanks. In this paper, we aim to incorporate social context and expert knowledge while focusing on learning representations of political actors to model their perspectives and facilitate downstream tasks.

\begin{figure*}
    \centering
    \includegraphics[width=1\linewidth]{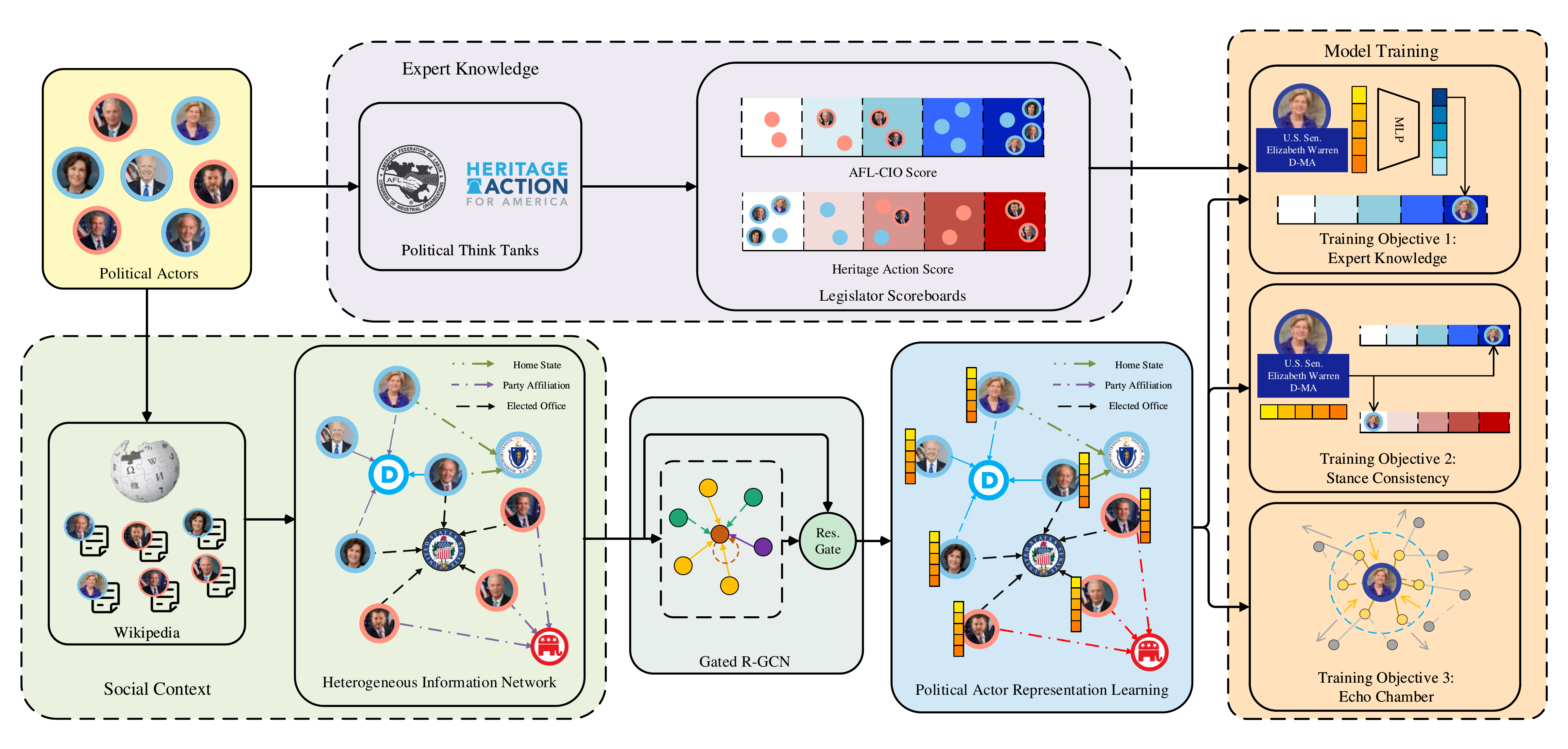}
        \caption{Overview of PAR, political actor representation learning with social context and expert knowledge.}
    \label{fig:overview}
\end{figure*}

\section{Methodology}
Figure \ref{fig:overview} presents an overview of \textbf{PAR} (\textbf{P}olitical \textbf{A}ctor \textbf{R}epresentation learning). We firstly collect a dataset of political actors from Wikipedia and political think tanks. We then construct a heterogeneous information network to jointly model political actors and social context information. Finally, we learn graph representations with gated relational graph convolutional networks (gated R-GCN) and train the framework with three different objectives to leverage expert knowledge and model various socio-political phenomena.

\subsection{Data Collection}
We firstly collect a dataset about political actors in the United States that were active in the past decade. For \textbf{social context} information, we retrieve the list of senators and congresspersons from the 114th congress to the 117th congress\footnote{https://www.congress.gov/}. We then retrieve their Wikipedia pages\footnote{https://github.com/goldsmith/Wikipedia} and extract these named entities: presidents, senators, congresspersons, governors, states, political parties, supreme court justices, government institutions, and office terms (117th congress etc.). In this way, we obtain 1,069 social and political entities. Based on these entities, we identify five types of relations: party affiliation, home state, political office, term in office, and appoint relationships. In this way, we obtain 9,248 heterogeneous edges. For \textbf{expert knowledge}, we use legislator scoreboards at AFL-CIO and Heritage Action, two political think tanks that lie in the opposite ends of the ideological spectrum. Specifically, we retrieve and extract each legislator's score in each office term. In this way, we obtain 777 scores from AFL-CIO and 679 scores from Heritage Action. We consolidate the collected social context and expert knowledge to serve as the dataset in our experiments. Although we focus on political actors in the United States in this paper, our data collection process is also applicable for other countries and time ranges.



\subsection{Graph Construction}
To better model the interactions between political entities and their shared social context, we propose to construct a heterogeneous information network (HIN) from the dataset. 

\subsubsection{Heterogeneous Nodes}
Based on the collected dataset, we select diversified entities that are essential factors in modeling the political process. Specifically, we use eight types of nodes to represent political actors and diversified social context entities.

\noindent \textit{\underline{$\mathcal{N}1$: Office Terms.}} We use four nodes to represent the 114th, 115th, 116th, 117th congress spanning from 2015 to 2022. We use these nodes to model different political scenarios and could be similarly extended to other time periods.

\noindent \textit{\underline{$\mathcal{N}2$: Legislators.}} We retrieve senators and congresspersons from the 114th to 117th congress and use one node to represent each distinct legislator.

\noindent \textit{\underline{$\mathcal{N}3$: Presidents.}} President is the highest elected office in the United States. We use three nodes to represent President Biden, Trump and Obama to match with the time range in $N1$.

\noindent \textit{\underline{$\mathcal{N}4$: Governors.}} State and local politics are also essential in analyzing the political process. We use one node to represent each distinct governor of 50 states within the time range of $N1$.

\noindent \textit{\underline{$\mathcal{N}5$: States.}} The home state of political actors is often an important indicator and helps connect different individuals. We use one node to represent each state in the United States.

\noindent \textit{\underline{$\mathcal{N}6$: Government Institutions.}} We use five nodes to represent the white house, senate, house of representatives, supreme court and governorship. These nodes enable our HIN to separate different political actors based on the office they hold.

\noindent \textit{\underline{$\mathcal{N}7$: Supreme Court Justices.}} Supreme court justices are nominated by presidents and approved by senators, which helps connect different types of political actors. We use one node to represent each supreme court justice within the time range of $N1$.

\noindent \textit{\underline{$\mathcal{N}8$: Political Parties.}} We use two nodes to represent the Republican Party and the Democratic Party in the United States.

For node features, we use pre-trained RoBERTa \citep{liu2019roberta} to encode the first paragraph of Wikipedia pages and average all tokens.

\subsubsection{Heterogeneous Relations}
\label{subsubsec:relation}
Based on $\mathcal{N}1$ to $\mathcal{N}8$, we extract five types of informative interactions between entities to complete the HIN structure. Specifically, we use five types of heterogeneous relations to connect different nodes and construct our political actor HIN.

\noindent \textit{\underline{$\mathcal{R}1$: Party Affiliation.}} We connect political actors and their affiliated political party with $R1$:
\begin{equation}
\mathcal{R}1 = (\mathcal{N}2 \cup \mathcal{N}3 \cup \mathcal{N}4) \times \mathcal{N}8
\end{equation}

\noindent \textit{\underline{$\mathcal{R}2$: Home State.}} We connect political actors with their home states with $R2$: 
\begin{equation}
\mathcal{R}2 = (\mathcal{N}2 \cup \mathcal{N}3 \cup \mathcal{N}4 \cup \mathcal{N}7) \times \mathcal{N}5
\end{equation}

\noindent \textit{\underline{$\mathcal{R}3$: Hold Office.}} We connect political actors with the political office they hold with $R3$:
\begin{equation}
\mathcal{R}3 = (\mathcal{N}2 \cup \mathcal{N}3 \cup \mathcal{N}4 \cup \mathcal{N}7) \times \mathcal{N}6
\end{equation}

\noindent \textit{\underline{$\mathcal{R}4$: Time in Office.}} If a political actor holds office during one of the time stamps in $N1$, we connect them with $R4$:
\begin{equation}
\mathcal{R}4 = (\mathcal{N}2 \cup \mathcal{N}3 \cup \mathcal{N}4 \cup \mathcal{N}7) \times \mathcal{N}1
\end{equation}

\noindent \textit{\underline{$\mathcal{R}5$: Appoint.}} Besides from being elected, certain political actors are appointed by others. We denote this relation with $R5$:
\begin{equation}
\mathcal{R}5 = (\mathcal{N}3 \times \mathcal{N}7) \cup (\mathcal{N}4 \times \mathcal{N}2)
\end{equation}


\subsection{Representation Learning}
Since nodes represent political actors, we learn node-level representations with gated R-GCN to jointly leverage social context and expert knowledge. Let $E = \{e_1, \cdot \cdot \cdot, e_n\}$ be $n$ entities in the HIN and $v_i$ be the initial features of entity $e_i$. Let $R$ be the heterogeneous relation set and $N_r(e_i)$ be entity $e_i$'s neighborhood with regard to relation type $r$. We firstly transform $v_i$ to serve as the input of graph neural networks:
\begin{equation}
    x_i^{(0)} = \phi(W_I \cdot v_i + b_I)
\end{equation}
where $\phi$ is leaky-relu, $W_I$ and $b_I$ are learnable parameters. We then propagate entity messages and aggregate them with gated R-GCN. For the $l$-th layer of gated R-GCN:
\begin{equation}
    u_i^{(l)} = \sum_{r \in R} \frac{1}{|N_r(e_i)|} \sum_{j \in N_r(e_i)}  f_r(x_j^{(l-1)}) + f_s(x_i^{(l-1)})
\end{equation}
where $f_s$ and $f_r$ are parameterized linear layers for self loops and edges of relation $r$, $u_i^{(l)}$ is the hidden representation for entity $e_i$ at layer $l$. We then calculate gate levels:
\begin{equation}
    g_i^{(l)} = \sigma(W_G \cdot [u_i^{(l)}, x_i^{(l-1)}] + b_G)
\end{equation}
where $W_G$ and $b_G$ are learnable parameters, $\sigma(\cdot)$ denotes the sigmoid function and $[\cdot , \cdot]$ denotes the concatenation operation. We then apply the gate mechanism to $u_i^{(l)}$ and $x_i^{(l-1)}$:
\begin{equation}
    x_i^{(l)} = \mathrm{tanh}(u_i^{(l)}) \odot g_i^{(l)} + x_i^{(l-1)} \odot (1 - g_i^{(l)})
\end{equation}
where $\odot$ is the Hadamard product. After $L$ layer(s) of gated R-GCN, we obtain node features $\{x^{(L)}_1,\cdots,x^{(L)}_n\}$ and the nodes representing political actors are extracted as learned representations.

\subsection{Model Training}
We propose to train PAR with a combination of three training objectives, which aligns representation learning with expert knowledge, ensures stance consistency and simulates the echo chamber phenomenon. The overall loss function of PAR is as follows:

\begin{equation}
    L = \lambda_1 L_1 + \lambda_2 L_2 + \lambda_3 L_3 + \lambda_4 \sum_{w \in \theta} ||w||_2^2
\label{equ:total_loss}
\end{equation}

\noindent where $\lambda_i$ is the weight of loss $L_i$ and $\theta$ are learnable parameters in PAR. We then present the motivation and detail of each loss function $L_1$, $L_2$ and $L_3$.

\subsubsection{Objective 1: Expert Knowledge}
The expert knowledge objective aims to align the learned representations with expert knowledge from political think tanks. We use the learned representations of political actors to predict their liberal and conservative stances, which are adapted from AFL-CIO and Heritage Action. Specifically:
\begin{equation}
\label{equ:lcmark}
\begin{aligned}
    l_i = \mathrm{softmax}(W_L \cdot x_i^{(L)} + b_L) \\
    c_i = \mathrm{softmax}(W_C \cdot x_i^{(L)} + b_C)
\end{aligned}
\end{equation}
where $l_i$ and $c_i$ are predicted stances towards liberal and conservative values, $W_L$, $b_L$, $W_C$ and $b_C$ are learnable parameters. Let $E_L$ and $E_C$ be the training set of AFL-CIO and Heritage Action scores, $\hat{l_i}$ and $\hat{c_i}$ denote the ground-truth among $D$ possible labels. We calculate the expert knowledge loss:

\begin{equation}
\label{equ:expert_loss}
    L_1 = - \sum_{e_i \in E_L} \sum_{d=1}^D \hat{l_{id}}\mathrm{log}(l_{id}) - \sum_{e_i \in E_C} \sum_{d=1}^D \hat{c_{id}}\mathrm{log}(c_{id})
\end{equation}

$L1$ enables PAR to align learned representations with expert knowledge from political think tanks.

\subsubsection{Objective 2: Stance Consistency}
The stance consistency objective is motivated by the fact that liberalism and conservatism are opposite ideologies, thus individuals often take inversely correlated stances towards them. We firstly speculate entities' stance towards the opposite ideology by taking the opposite of the predicted stance:
\begin{equation}
    \Tilde{l_i} = \psi(D - \mathrm{argmax}(c_i)), \ \ \Tilde{c_i} = \psi(D - \mathrm{argmax}(l_i))
\end{equation}
where $\psi$ is the one-hot encoder, $\mathrm{argmax}(\cdot)$ calculates the vector index with the largest value, $D$ is the number of stance labels, $\Tilde{l_i}$ and $\Tilde{c_i}$ are labels derived with stance consistency. We calculate the loss function $L_2$ measuring stance consistency:
\begin{equation}
    L_2 = - \sum_{e_i \in E} \ \sum_{d=1}^D \ (\Tilde{l_{id}} \ \mathrm{log}(l_{id}) \ + \ \Tilde{c_{id}} \ \mathrm{log}(c_{id}) )
\end{equation}
where $E = E_L \cap E_C$. As a result, the loss function $L_2$ enables PAR to ensure ideological stance consistency among political actors.

\subsubsection{Objective 3: Echo Chamber}
The echo chamber objective is motivated by the echo chamber phenomenon \citep{jamieson2008echo, barbera2015tweeting}, where social entities tend to reinforce their narratives by forming small and closely connected interaction circles. We simulate echo chambers by assuming that neighboring nodes in the HIN have similar representations while non-neighboring nodes have different representations. We firstly define the positive and negative neighborhood of entity $e_i$:
\begin{equation}
\begin{aligned}
P_{e_i} = \{e \ \ | \ \exists \ r \in R \ \ s.t. \ \ e \in N_r(e_i) \} \\
N_{e_i} = \{e \ \ | \ \forall \ r \in R \ \ s.t. \ \ e \notin N_r(e_i) \}
\end{aligned}
\end{equation}

We then calculate the echo chamber loss:
\begin{equation}
\begin{aligned}
    L_3 = - \sum_{e_i \in E} \sum_{e_j \in P_{e_i}} \mathrm{log}(\sigma(x_i^{{(L)^T}} x_j^{(L)})) \\
    + Q \cdot \sum_{e_i \in E} \sum_{e_j \in N_{e_i}} \mathrm{log}(\sigma(-x_i^{{(L)^T}} x_j^{(L)}))
\end{aligned}
\label{equ:unsupervised_loss}
\end{equation}
where $Q$ denotes the weight for negative samples. $L_3$ enables PAR to model the echo chamber phenomenon in real-world socio-economic systems.

\section{Experiment}
After learning political actor representations with PAR, we study whether they are effective in computational political science and reveal real-world political behavior. Specifically, we test out PAR on political perspective detection and roll call vote prediction, two political text understanding tasks that emphasizes political actor modeling. We leverage PAR in these tasks to examine whether it could contribute to the better understanding of political text. We then study the learned representations of PAR and whether they provide new insights into real-world politics.

\subsection{Political Perspective Detection}
Political perspective detection aims to detect stances in text such as public statements and news articles, which generally mention many political actors to provide context and present arguments. Existing methods model political actors with masked entity models \citep{li2021using} and knowledge graph embedding techniques \citep{feng2021knowledge}. We examine whether PAR is more effective than these models and consequently improve political perspective detection.

\begin{table}[t]
    \centering
    \resizebox{\linewidth}{!}{
        \begin{tabular}{l l|c c|c c}
             \toprule[1.5pt] \multirow{2}{*}{\textbf{Method}} & \multirow{2}{*}{\textbf{Setting}} & \multicolumn{2}{c|}{\textbf{SemEval}} & \multicolumn{2}{c}{\textbf{AllSides}} \\ 
             & & \textbf{Acc} & \textbf{MaF} & \textbf{Acc} & \textbf{MaF} \\ \midrule[0.75pt]
             \multirow{2}{*}{\textbf{CNN}}&GloVe & $79.63$ & $N/A$ & $N/A$ & $N/A$ \\
             &ELMo & $84.04$ & $N/A$  & $N/A$  & $N/A$  \\ \midrule[0.75pt]
             \multirow{4}{*}{\textbf{HLSTM}}&GloVe & $81.58$ & $N/A$  & $N/A$  & $N/A$  \\
             &ELMo & $83.28$ & $N/A$  & $N/A$  & $N/A$  \\
             &Embed & $81.71$ & $N/A$  & $76.45$ & $74.95$ \\
             &Output & $81.25$ & $N/A$  & $76.66$ & $75.39$ \\ \midrule[0.75pt]
             \textbf{BERT} & base & 84.03 & 82.60 & 81.55 & 80.13 \\ \midrule[0.75pt]
             \multirow{3}{*}{\textbf{MAN}}&GloVe & $81.58$ & $79.29$ & $78.29$ & $76.96$ \\
             &ELMo & $84.66$ & $83.09$ & $81.41$ & $80.44$ \\
             &Ensemble & $86.21$ & $84.33$ & $85.00$ & $84.25$ \\ \midrule[0.75pt]
             \multirow{5}{*}{\textbf{KGAP}} 
             & TransE & $89.56$ & $84.94$ & $86.02$ & $85.52$ \\
             & TransR & $88.54$ & $83.45$ & $85.15$ & $84.61$ \\
             & DistMult & $88.51$ & $83.63$ & $84.47$ & $83.90$ \\
             & HolE & $88.85$ & $83.68$ & $84.78$ & $84.24$ \\
             & RotatE & $88.84$ & $84.04$ & $85.61$ & $85.11$ \\ \midrule[0.75pt]
             \textbf{KGAP} & \textbf{PAR} & $\textbf{91.30}$ & $\textbf{87.78}$ & $\textbf{86.81}$ & $\textbf{86.33}$ \\
             \bottomrule[1.5pt]
        \end{tabular}
    }
    \caption{Political perspective detection performance on two benchmark datasets. Acc and MaF denote accuracy and macro-averaged F1-score. N/A indicates that the result is not reported in previous works.}
    \label{tab:perspective}
\end{table}

\subsubsection{Datasets}
We follow previous works \citep{li2021using, feng2021knowledge} and adopt two political perspective detection benchmarks: SemEval and Allsides. SemEval \citep{SemEval} aims to identify whether a news article follows hyperpartisan argumentation. We follow the 10-fold cross validation setting and the exact same folds established in \citet{li2021using} so that the results are directly comparable. Allsides \citep{li2019encoding} provides left, center, or right labels for three-way classification. We follow the 3-fold cross validation setting and the exact same folds established in \citet{li2019encoding} so that the results are directly comparable.

\subsubsection{Baselines}
We compare PAR with competitive baselines:
\begin{itemize}[leftmargin=*]
    \item \textbf{CNN} \citep{CNNglove} achieves the best performance in the SemEval 2019 Task 4 contest \citep{SemEval}. It uses convolutional neural networks along with word embeddings \textbf{GloVe} \citep{CNNglove} and \textbf{ELMo} \citep{ELMo} for perspective detection.
    \item \textbf{HLSTM} \citep{HLSTM} is short for hierarchical long short-term memory networks. \citet{li2019encoding} combines it with \textbf{GloVe} and \textbf{ELMo} word embeddings. \citet{li2021using} leverages Wikipedia2Vec \citep{yamada2018wikipedia2vec} and masked entity models while using different concatenation strategies (\textbf{HLSTM\_Embed} and \textbf{HLSTM\_Output}).
    \item \textbf{BERT} \citep{devlin2018bert} is fine-tuned on the task of political perspective detection.
    \item \textbf{MAN} \citep{li2021using} leverages social and linguistic information for pre-training a BERT-based model and fine-tune on the task of political perspective detection.
    \item \textbf{KGAP} \citep{feng2021knowledge} models legislators with RoBERTa and knowledge graph embeddings: \textbf{TransE} \citep{TransE}, \textbf{TransR} \citep{transr}, \textbf{DistMult} \citep{distmult}, \textbf{HolE} \citep{hole}, and \textbf{RotatE} \citep{sun2019rotate}. It then constructs document graphs with text and legislators as nodes and uses GNNs for stance detection.
    \item \textbf{KGAP\_PAR}: To examine whether PAR is effective in political perspective detection, we replace the knowledge graph embeddings in \textbf{KGAP} with political actor representations learned with PAR.
\end{itemize}

\subsubsection{Results}
We evaluate PAR and baselines and present model performance in Table \ref{tab:perspective}, which demonstrates that PAR achieves state-of-the-art performance on both datasets. The fact that PAR outperforms MAN and KCD, two methods that also take political actors into account, shows that PAR learns high-quality representations that results in performance gains.

\begin{table}[]
    \centering
    \resizebox{0.8\linewidth}{!}{
    \begin{tabular}{c|c c}
         \toprule[1.5pt] \multirow{2}{*}{\textbf{Method}} & \multicolumn{2}{c}{\textbf{Setting}} \\ 
         & \textbf{random} & \textbf{time-based} \\ \midrule[1pt]
         \textbf{majority} & $77.48$ & $77.40$ \\
         \textbf{ideal-point-wf} & $85.37$ & $N/A$ \\
         \textbf{ideal-point-tfidf} & $86.48$ & $N/A$ \\
         \textbf{ideal-vector} & $87.35$ & $N/A$ \\
         \textbf{CNN} & $87.28$ & $81.97$ \\
         \textbf{CNN+meta} & $88.02$ & $84.30$ \\
         \textbf{LSTM+GCN} & $88.41$ & $85.82$ \\
         \textbf{Vote} & $90.22$ & $89.76$ \\
         \textbf{RoBERTa} & $87.59$ & $87.56$ \\
         \textbf{TransE} & $82.70$ & $80.06$ \\
         \textbf{PAR} & \textbf{90.33} & \textbf{89.92} \\ \bottomrule[1.5pt]
    \end{tabular}
    }
    \caption{Roll call vote prediction performance (accuracy) with two experiment settings. N/A indicates that the result is not reported in previous works.}
    \label{tab:vote}
\end{table}

\begin{figure*}[]
    \centering
    \includegraphics[width=1\linewidth]{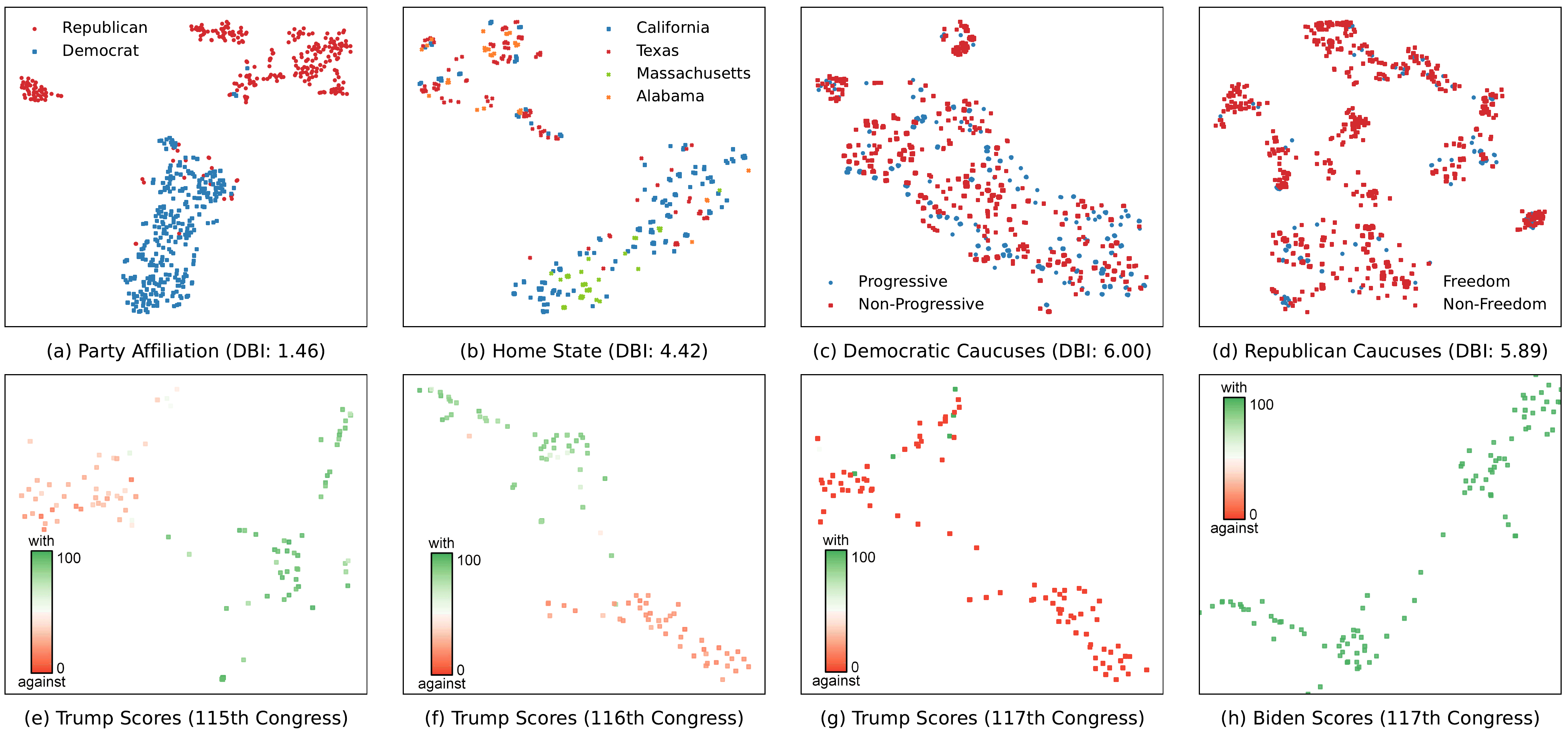}
    \caption{Using t-sne to visualize learned representations of political actors. DBI denotes the Davies-Bouldin Index.}
    \label{fig:replearn}
\end{figure*}

\subsection{Roll Call Vote Prediction}
Roll call vote prediction aims to predict whether a legislator will vote in favor or against a specific legislation. Ideal points \citep{gerrish2011predicting}, ideal vectors \citep{kraft2016embedding}, and \citet{mou2021align} have been proposed to model ideological preferences. We examine whether PAR is more effective than these models in roll call vote prediction.

\subsubsection{Datasets}
We adopt the datasets and settings proposed in \citet{mou2021align} to evaluate PAR and competitive baselines. Specifically, for the random setting, voting records in the 114th and 115th congresses are randomly split into 6:2:2 for training, validation, and testing. For the time-based setting, the voting records in the 114th congress is split into 8:2 for training and validation, while the voting records in the 115th congress is used for testing.

\subsubsection{Baselines}
We compare PAR with competitive baselines:
\begin{itemize}[leftmargin=*]
    \item \textbf{majority} predicts all votes as \textit{yea}.
    \item \textbf{ideal-point-wf} and \textbf{ideal-point-tfidf} \citep{gerrish2011predicting} adopt word frequency and TFIDF of legislation text as features and leverage the ideal point model for vote prediction.
    \item \textbf{ideal-vector} \citep{kraft2016embedding} learns distributed representations with legislator and bill text.
    \item \textbf{CNN} \citep{kornilova2018party} encodes bill text with CNNs. On top of that, \textbf{CNN+meta} introduces metadata of bill sponsorship to the process.
    \item \textbf{LSTM+GCN} \citep{yang2021joint} leverages LSTM and GCN to jointly update representations of legislations and legislators.
    \item \textbf{Vote} \citep{mou2021align} proposes to align statements on social networks with voting records.
    \item \textbf{RoBERTa} and \textbf{TransE} use RoBERTa \citep{liu2019roberta} encoding of legislator description on Wikipedia or TransE \citep{TransE} embeddings for legislator representation learning. They are then concatenated with RoBERTa encoded bill text for vote prediction.
    \item \textbf{PAR} concatenates political actor representations learned with PAR and RoBERTa encoded legislator text for roll call vote prediction.
\end{itemize}

\subsubsection{Results}
We evaluate PAR and competitive baselines on roll call vote prediction and present model performance in Table \ref{tab:vote}. PAR achieves state-of-the-art performance, outperforming existing baselines that model political actors in different ways. As a result, PAR learns high-quality representations of political actors that provide political knowledge and augment the vote prediction process.

\begin{figure*}
    \centering
    \includegraphics[width=1\linewidth]{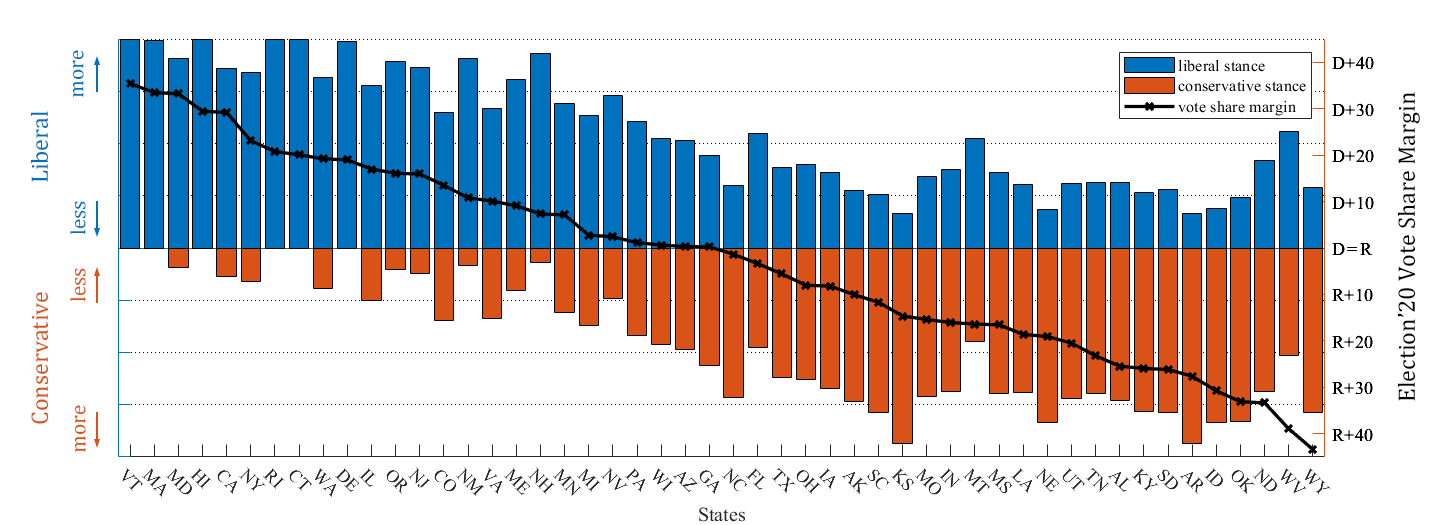}
    \caption{State stances predicted by PAR compared to vote shares in the 2020 U.S. presidential election.}
    \label{fig:state_stance}
\end{figure*}

\subsection{Political Findings of PAR}
PAR learns political actor representations with social context and expert knowledge, which has been proven effective in political perspective detection and roll call vote prediction. We further examine whether PAR provides new insights into the political behavior of legislators, states, and governors.

\subsubsection{Legislators and PAR}
To examine whether legislator representations learned with PAR align well with different social and political factors, we adopt t-sne \citep{maaten_visualizing_2008} to illustrate our learned representations of political actors in Figure \ref{fig:replearn}.

\noindent \textbf{Social context.}
Figure \ref{fig:replearn} (a) and (b) illustrate the correlation between learned representations and social context factors such as party affiliation and home state. It is illustrated that legislators from the same party and the same home state tends to be similar in the representation space. In addition, we calculate the DBI scores \citep{DBIscore} to quantitatively analyze the collocation. In conclusion, the learned representations successfully reflect these social context information.

\noindent \textbf{Congressional caucus.} 
Figure \ref{fig:replearn} (c) and (d) demonstrate the correlation between learned representations and major congressional caucuses in both parties. The progressive caucus in the democratic party and the freedom caucus in the republican party are often viewed as more ideological wings of the party. Both the illustration and the DBI scores suggest little collocation among different caucuses. As a result, PAR reveals that despite widening ideological gaps between party factions \citep{cohen2016party}, inter-party differences still outweigh intra-party differences in contemporary U.S. politics.

\noindent \textbf{Voting records.} 
Figure \ref{fig:replearn} (e), (f), (g) and (h) illustrate how often does a legislator vote for or against the sitting president. We retrieve this information from FiveThirtyEight\footnote{\url{https://fivethirtyeight.com/}} and illustrate voting records with color gradients. "Trump scores" and "Biden scores" indicate what percentage out of all votes did a legislator vote with or against the official stance of the president. As a result, our learned representations of legislators in the 115th and 116th congress correlate well with their voting records, while the 117th congress might have not hold enough votes for an accurate categorization\footnote{Voting records of the 117th congress is collected in June 2021, only five months into its term.}.

\subsubsection{States and PAR}
Political commentary often uses "blue state", "red state", and "swing state" to describe the ideological preference of states in the U.S. We examine the ideological scores of states learned by PAR ($l_i$ and $c_i$ in Equation (\ref{equ:lcmark})) and compare them with results in the 2020 U.S. presidential election in Figure \ref{fig:state_stance}. It is illustrated that PAR stance predictions highly correlate with the 2020 presidential election results. In addition, PAR reveals that Pennsylvania and North Carolina, two traditional swing states, are actually more partisan than expected. PAR also suggests that Georgia is the most electorally competitive state in the United States and we will continue to monitor this conclusion in future elections.

\subsubsection{Governors and PAR}
\begin{figure}
    \centering
    \includegraphics[width=1\linewidth]{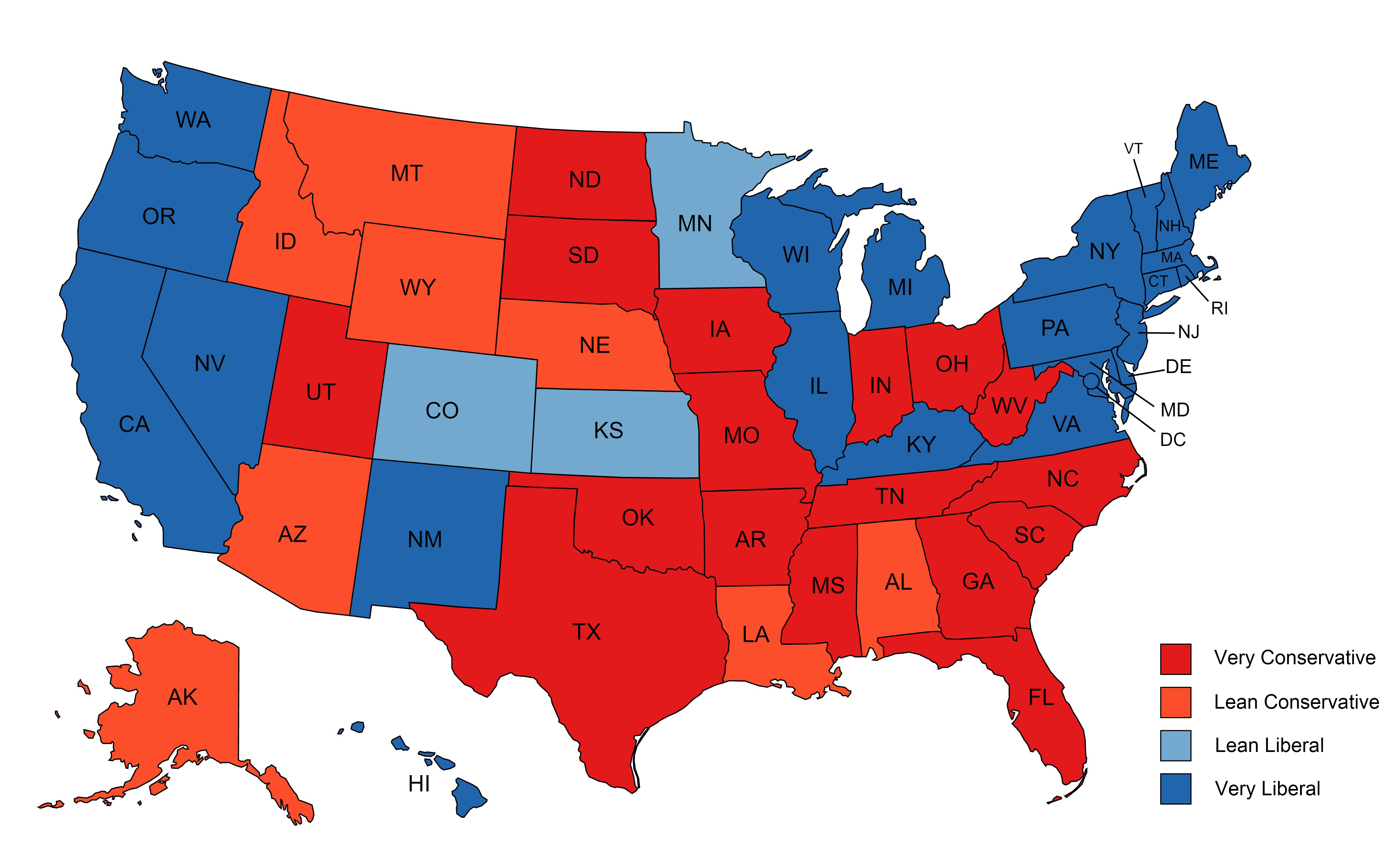}
    \caption{Governor stances predicted by PAR.}
    \label{fig:governor_stance}
\end{figure}

Political experts typically study and evaluate the stances of presidents and federal legislators, while state-level officials such as governors are also essential in governance and policy making \citep{beyle1988governor}. PAR complements the understanding of state-level politics by learning ideological scores for governors. Specifically, we use $l_i$ and $c_i$ in Equation (\ref{equ:lcmark}) to evaluate the ideological position of governors and present PAR's predictions of governor stances in all 50 U.S. states in Figure \ref{fig:governor_stance}\footnote{This figure is created with the help of \url{mapchart.net}.}. It is no surprise that governors from partisan strongholds such as California and Utah hold firm stances. Conventional wisdom often assumes that in order to win "swing states" or electorally challenging races, one has to compromise their ideological stances and lean to the center. However, PAR reveals exceptions to this rule, such as Andy Beshear (D-KY) and Ron DeSantis (R-FL), which is also suggested in political commentary \citep{andybeshear, rondesantis}.

\section{Conclusion}
In this paper, we present PAR, a framework to learn representations of political actors with social context and expert knowledge. Specifically, we retrieve social context information from Wikipedia and expert knowledge from political think tanks, construct a HIN to model legislators and learn representations with gated R-GCNs and three training objectives. Extensive experiments demonstrate that PAR advances the state-of-the-art in political perspective detection and roll call vote prediction. PAR further provides novel insights into political actors through its representation learning process.

\section{Limitations}

\begin{itemize}[leftmargin=*]
    \item PAR proposes to learn representations for political actors with the help of social context and expert knowledge. Though it achieved great performance on several tasks and provided interesting political insights, it might reinforce political stereotypes, such as people from "red states" are often assumed to be more conservative. We leave for future work how to mitigate the potential bias in political actor representations.
    \item In this paper, we focus on political actors in the United States instead of other countries. However, our proposed framework could be easily extended to other scenarios by leveraging the Wikipedia pages of political actors for social context and political think tanks in other countries for expert knowledge.
\end{itemize}

\section*{Acknowledgements}
This work was supported by the National Key Research and Development Program of China (No. 2020AAA0108800), National Nature Science Foundation of China (No. 62192781, No. 62272374, No. 61872287, No. 62250009, No. 62137002), Innovative Research Group of the National Natural Science Foundation of China (61721002), Innovation Research Team of Ministry of Education (IRT\_17R86), Project of China Knowledge Center for Engineering Science and Technology and Project of Chinese academy of engineering ``The Online and Offline Mixed Educational Service System for ‘The Belt and Road’ Training in MOOC China''.

We would like to thank the reviewers and area chair for their constructive feedback. We would also like to thank all LUD lab members for our collaborative research environment. 

\bibliography{custom}
\bibliographystyle{acl_natbib}

\clearpage

\appendix

\section{Expert Knowledge Prediction}
We propose to train PAR with three objectives, the first one $L_1$ being expert knowledge prediction, where the model learns to predict how liberal or conservative a given legislator is. To examine whether the PAR architecture successfully conducts expert knowledge prediction, we compare PAR with various text~\citep{pedregosa2011scikit,MEANbiasfeature,pennington2014glove,liu2019roberta,beltagy2020longformer} and graph~\citep{GCN,GAT,SAGE,TransformerConv,ResGatedGraphConv} analysis baselines. Table \ref{tab:big} presents model performance on the expert knowledge task. It is demonstrated that PAR achieves the best performance compared to various text and graph analysis baselines, suggesting our learned representations successfully reflect expert knowledge from political think tanks. Apart from that, graph-based models generally outperform text-based methods, which suggests that the constructed HIN is essential in modeling political actors.

\begin{table}[t]
    \centering
    \resizebox{\linewidth}{!}{
    \begin{tabular}{c|c c|c c c}
         \toprule[1.5pt] \textbf{Method} & \textbf{Text} & \textbf{Graph} & \textbf{Acc} & \textbf{MaF} & \textbf{MiF} \\ \midrule[0.75pt]
         Linear BoW & \checkmark &  & $68.49$&	$40.00$&	$68.53$ \\
         Bias Features & \checkmark & & $47.26$&	$20.08$&	$47.10$ \\
         Glove & \checkmark & & $52.05$&	$26.94$&	$52.01$ \\
         RoBERTa & \checkmark & & $71.92$&	$49.70$&	$71.87$ \\
         LongFormer & \checkmark & & $68.49$&	$42.27$&	$68.56$ \\ \midrule[0.75pt]
         GCN & \checkmark & \checkmark & $74.66$&	$54.16$&	$74.46$ \\
         GAT & \checkmark & \checkmark & $78.08$&	$55.82$&	$78.17$ \\
         GraphSAGE & \checkmark & \checkmark & $75.34$&	$51.39$&	$75.43$ \\
         TransformerConv & \checkmark & \checkmark & $77.40$&	$55.63$&	$77.48$ \\
         ResGatedConv & \checkmark & \checkmark & $76.03$&	$54.31$&	$75.97$ \\ \midrule[0.75pt]
         \textbf{Ours} & \checkmark & \checkmark &  $\textbf{80.82}$&	$\textbf{60.37}$&	$\textbf{80.89}$ \\ \bottomrule[1.5pt]
    \end{tabular}
    }
    \caption{Our model's performance on the expert knowledge prediction task compared to various text and graph analysis baselines. Acc, MaF and MiF denote accuracy, macro and micro-averaged F1-score.}
    \label{tab:big}
\end{table}

\section{Ablation Study}
PAR aims to learn representations of political actors with social context and expert knowledge as well as three training objectives. We conduct ablation study to examine their effect in the representation learning process. We report model performance on the expert knowledge prediction task and follow the same dataset splits as in Table \ref{tab:big}.

\subsection{Social Context}
\begin{figure}[t]
    \centering
    \includegraphics[width=0.9\linewidth]{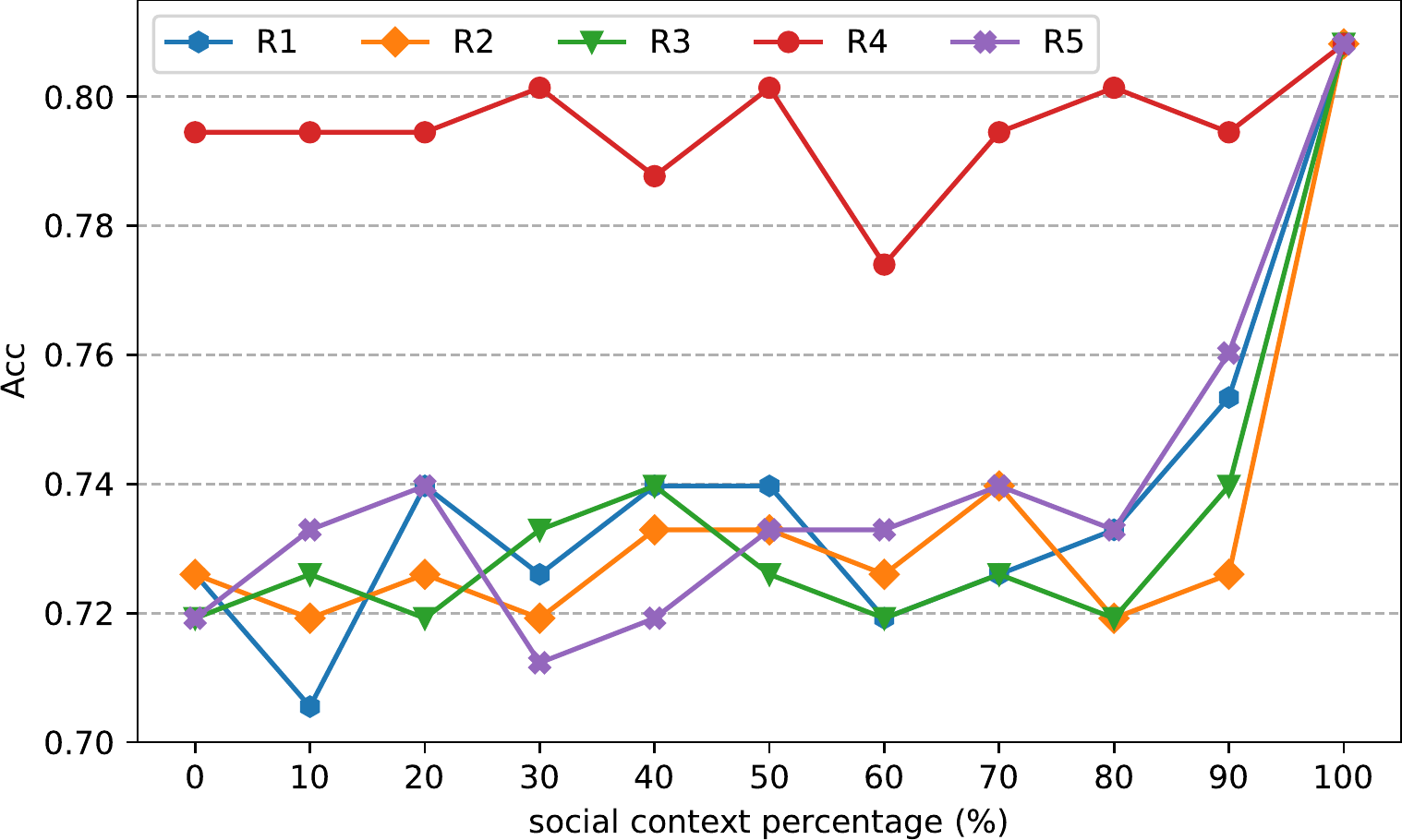}
    \caption{Ablation study of social context information. $R1$ to $R5$ follows that in Section \ref{subsubsec:relation}.}
    \label{fig:ablation_social_context}
\end{figure}

We use five types of heterogeneous relations $R1$ to $R5$ to connect different entities based on their social context. We randomly and gradually remove five types of social context edges in the constructed HIN and report model performance in Figure \ref{fig:ablation_social_context}. It is illustrated that all relations but $R4$ (Time in Office) significantly contributes to the overall performance. Besides, Figure \ref{fig:ablation_social_context} illustrates a great gap between $90\%$ and $100\%$ edges, suggesting the importance of a complete HIN structure.

\subsection{Expert Knowledge}
\begin{figure}[t]
    \centering
    \includegraphics[width=0.9\linewidth]{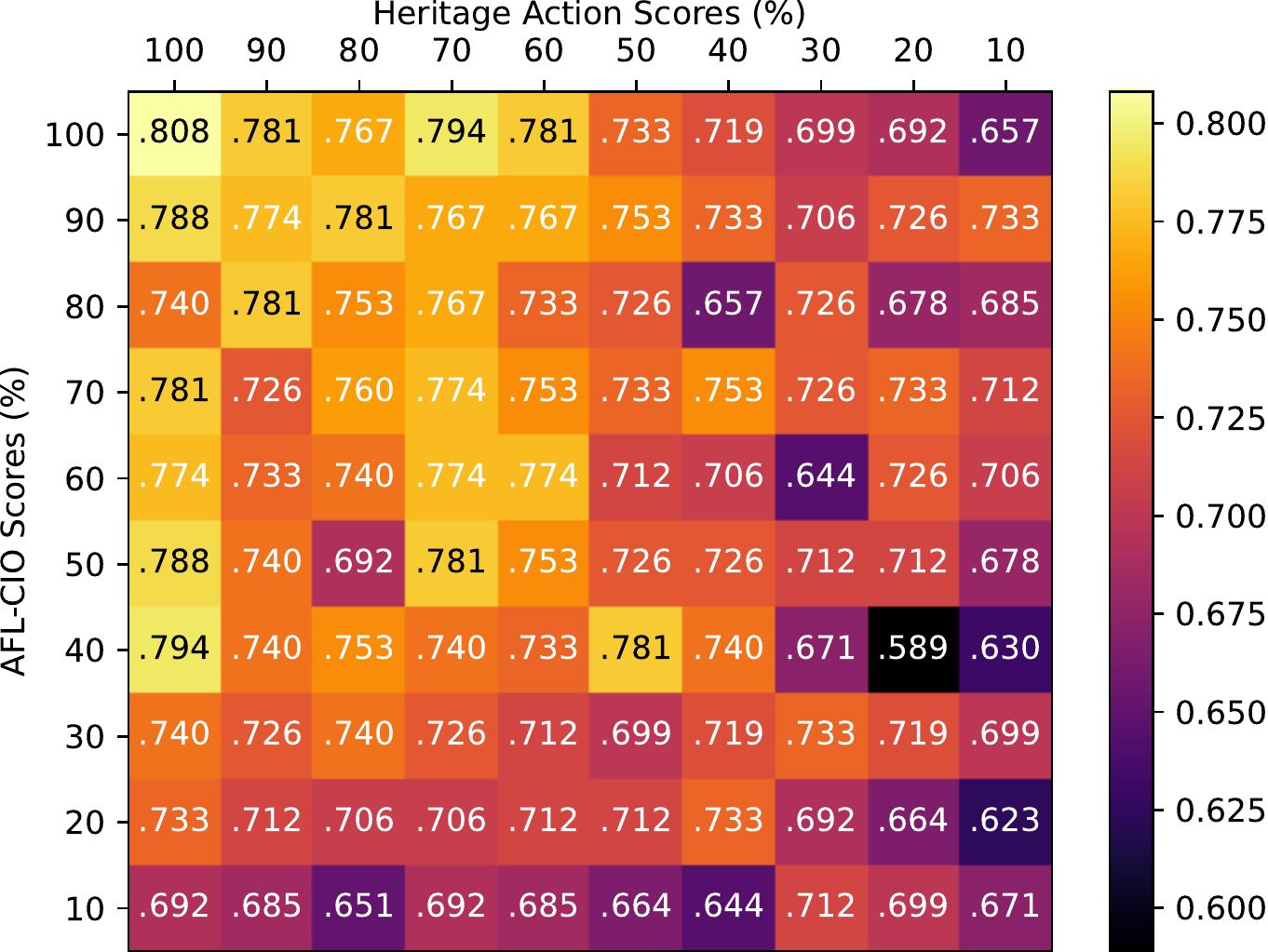}
    \caption{Ablation study of expert knowledge. We report accuracy on the expert knowledge prediction task.}
    \label{fig:ablation_expert_knowledge}
\end{figure}

We learn legislator representations with the help of two political think tanks: AFL-CIO and Heritage Action. We retrieve their evaluation of political actors amd construct the expert knowledge objective for training. To examine the effect of expert knowledge in our proposed approach, we gradually remove expert knowledge labels in $L_1$ and report model performance in Figure \ref{fig:ablation_expert_knowledge}. It is illustrated that our performance drops with partial expert knowledge from either AFL-CIO or Heritage Action, which indicates that expert knowledge is essential in the representation learning process.

\subsection{Training Objectives}
\begin{figure}[t]
    \centering
    \includegraphics[width=0.9\linewidth]{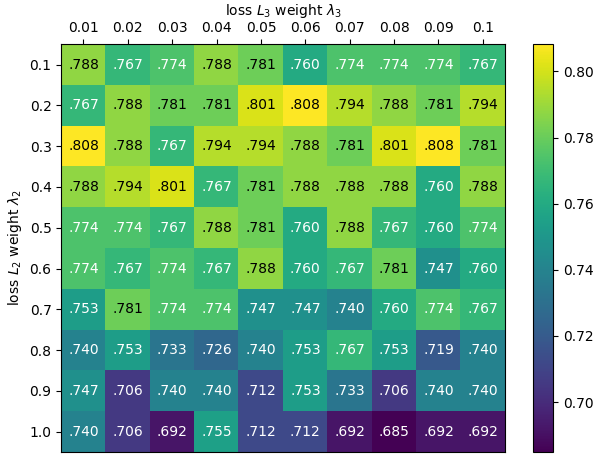}
    \caption{Model accuracy with different loss weights.}
    \label{fig:loss_weight}
\end{figure}

We propose to train our framework with three objectives. To examine their effect, we train our method with different combinations of $L_1$, $L_2$ and $L_3$ and report performance in Table \ref{tab:ablation_training_objectives}. Our model performs best with all three training objectives, proving the effectiveness of our loss function design. We further study the influence of loss weights $\lambda_1$, $\lambda_2$, $\lambda_3$ and $\lambda_4$. We fix $\lambda_1 = 1$ and $\lambda_4 = 10^{-5}$, present model performance under different settings of loss weights for auxiliary tasks $\lambda_2$ and $\lambda_3$ in Figure \ref{fig:loss_weight}. It is illustrated that $0.2 \leq \lambda_2 \leq 0.3$ and $0.01 \leq \lambda_3 \leq 0.1$ would generally lead to an effective balance of three different training objectives.

\begin{table}[t]
    \centering
    \resizebox{0.9\linewidth}{!}{
    \begin{tabular}{c|c c c}
         \toprule[1.5pt] \textbf{Loss Function(s)} & \textbf{Acc} & \textbf{MaF} & \textbf{MiF} \\ \midrule[0.75pt]
         $L_1$ only & $78.08$ & $56.43$ & $78.17$ \\
         $L_1$ and $L_2$ & $79.45$ & $55.97$ & $79.52$ \\
         $L_1$ and $L_3$ & $76.03$ & $51.91$ & $76.11$ \\
         $L_1$, $L_2$, and $L_3$ & $\textbf{80.82}$ & $\textbf{60.37}$ & $\textbf{80.89}$ \\ \bottomrule[1.5pt]
    \end{tabular}
    }
    \caption{Ablation study of three training objectives.}
    \label{tab:ablation_training_objectives}
\end{table}

\subsection{Graph Learning}
We adopted five relations $R1$ to $R5$ to connect eight types of entities $N1$ to $N8$, so that the graph is heterogeneous. To examine whether the graph heterogeneity contributes to model performance, we substitute gated R-GCNs with homogeneous GNNs such as GCN, GAT, and GraphSAGE. Table \ref{tab:gnn_operator} shows mixed results, where PAR performs best with gated R-GCNs while R-GCN does not outperform GraphSAGE.

\begin{table}[t]
    \centering
    \resizebox{0.9\linewidth}{!}{
    \begin{tabular}{c c|c c c}
         \toprule[1.5pt] \textbf{GNN operator} & \textbf{Het.} & \textbf{Acc} & \textbf{MaF} & \textbf{MiF}  \\ \midrule[0.75pt]
         GCN & \xmark & 	$76.03$ &	$58.31$ &	$78.08$ \\
         GAT & \xmark &	$77.40$ &	$59.01$ &	$78.85$ \\
         SAGE & \xmark & $78.77$ &	$58.04$ & 	$78.81$ \\
         R-GCN & \checkmark & $78.08$ &	$55.61$&	$78.15$ \\
         Gated R-GCN & \checkmark & $\textbf{80.82}$ & $\textbf{60.37}$ & $\textbf{80.89}$ \\ \bottomrule[1.5pt]
    \end{tabular}
    }
    \caption{Model performance with different GNNs. We adopt gated R-GCN and achieves the best performance. Het. denotes whether it supports heterogeneous graphs.}
    \label{tab:gnn_operator}
\end{table}

\section{Error Analysis}
We manually examined the part of the results in the roll call vote prediction task. Among the (legislator, bill) pairs where PAR made a wrong prediction, it is often the case that the legislator has voted across party lines. This suggests that more information about these legislators are required to achieve more fine-grained analysis on these borderline cases.

\section{Reproducibility Details}
In this section, we provide additional details to facilitate reproducing our results and findings. We submit data and code as supplementary material and commit to make them publicly available upon acceptance to facilitate reproduction.

\subsection{Hyperparameters}
We present the hyperparameter settings of our proposed approach in Table \ref{tab:hyperparameter}. These hyperparameters are manually tuned. We follow these settings throughout the paper unless stated otherwise.

\begin{table}[t]
    \centering
    \resizebox{1\linewidth}{!}{
    \begin{tabular}{c|c|c|c}
         \toprule[1.5pt] 
         \textbf{Hyperparameter} & \textbf{Value} & \textbf{Hyperparameter} & \textbf{Value}\\
         \midrule[0.75pt]
         RoBERTa size & $768$ & GNN size & $512$ \\ 
         optimizer & Adam & learning rate & $1e-3$ \\
         batch size & $64$ & max epochs & $100$ \\
         $L$ & $2$ & $\phi$ & ReLU \\
         $Q$ & $-0.1$ & \#negative sample & $2$ \\
         $\lambda_1$ & $0.01$ & $\lambda_2$ & $0.2$ \\
         $\lambda_3$ & $1$ & $\lambda_4$ & $1e-5$ \\ \bottomrule[1.5pt]
    \end{tabular}
    }
    \caption{Hyperparameters of our proposed approach.}
    \label{tab:hyperparameter}
\end{table}

\subsection{Implementation}
We use pytorch~\citep{paszke2019pytorch}, pytorch lightning~\citep{Falcon_PyTorch_Lightning_2019}, torch geometric~\citep{torchgeometric}, and the transformers library~\citep{wolf-etal-2020-transformers} for implementation. All implemented codes are available as supplementary material.

\subsection{Computation}
Our proposed approach has a total of 2.0M learnable parameters with hyperparameters in Table \ref{tab:hyperparameter}. We train it on a Titan X GPU with 12GB memory. It takes approximately 0.6 GPU hours for training with hyperparameters in Table \ref{tab:hyperparameter}.

\section{Experiment Details}

\subsection{Political Perspective Detection}
To examine whether our learned representations of political actors would benefit perspective analysis, we replace TransE in \citet{feng2021knowledge} with our learned representations. We use the GRGCN setting in \citet{feng2021knowledge} as model backbone. We maintain the same evaluation settings to ensure a fair comparison and highlight the effectiveness of our learned representations compared to TransE.

\subsection{Roll Call Vote Prediction}
We make our best effort to maintain the same experiment settings as \citet{mou2021align} while their might be minor differences. For the \textit{random} setting, we conduct roll call vote prediction for legislators in the 114th and 115th congress. We follow the same 6:2:2 split. For the \textit{time-based} setting, we use the 114th congress as training and validation set and the 115th congress as test set.

\subsection{Expert Knowledge Prediction}
\label{subsec:expertknowledgeprediction}
We collect expert knowledge about legislators from two political think tanks, which assign a continuous score $s$ from 0 to 1 to indicate how well a political actor aligns with their agenda. We construct a classification task from expert knowledge by creating five labels: strongly favor ($0.9 \le s \leq 1$), favor ($0.75 \le s \leq 0.9$), neutral ($0.25 \le s \leq 0.75$), oppose ($0.1 \le s \leq 0.25$), and strongly oppose ($0 \leq s \leq 0.1$). In this way, we adapt from expert knowledge to derive liberal and conservative labels for legislators and thus $D=5$ for Equation \ref{equ:expert_loss}. We use 7:2:1 to partition them into training, validation and test sets. We calculate evaluation metrics on the liberal and conservative set separately, and present the harmonic mean of metrics. In this way, the presented results accurately and comprehensively reflect how our approach and baselines perform on both political think tanks. For text-based baselines, we encode Wikipedia summaries of entities with these methods and predict their stances with two fully connected layers. For graph-based baselines, we train them with the constructed HIN and the expert knowledge training objective.

\subsection{Figure \ref{fig:state_stance}}
As detailed in Section \ref{subsec:expertknowledgeprediction}, each entity in PAR's HIN has two set of scores: liberal scores $(l_0, l_1, l_2, l_3, l_4)$ and conservative scores $(c_0, c_1, c_2, c_3, c_4)$, denoting the probability that the entity is strongly against, against, neutral, favor, or strongly favor liberal/conservative values. We use $\sum_{i=0}^4 i \cdot l_i$ and $\sum_{i=0}^4 i \cdot c_i$ to obtain continuous values of state stances and let them be the height of blue and red bars in Figure \ref{fig:state_stance}.

\subsection{Figure \ref{fig:governor_stance}}
We use the liberal scores of governors learned by PAR $(l_0, l_1, l_2, l_3, l_4)$ to infer their ideological stances. Specifically, we use "very conservative", "lean conservative", "lean liberal", and "very liberal" to denote that $\mathrm{argmax} (l_0, l_1, l_2, l_3, l_4) = 0, 1, 3, 4$ respectively. In addition, there are no governor that has $\mathrm{argmax} (l_0, l_1, l_2, l_3, l_4) = 2$ so "neutral" is omitted from the figure.

\end{document}